\documentclass{article}

\usepackage{arxiv}

\usepackage[utf8]{inputenc} 
\usepackage[T1]{fontenc}    
\usepackage{hyperref}       
\usepackage{url}            
\usepackage{booktabs}       
\usepackage{amsfonts}       
\usepackage{nicefrac}       
\usepackage{microtype}      
\usepackage{graphicx}

\usepackage{biblatex}
\usepackage[dvipsnames]{xcolor}

\title{A small note on variation in segmentation annotations}

\author{
  Silas~Nyboe~Ørting\\
  Center for Quantification of Imaging data from MAX IV (QIM)\\
  Department of Computer Science\\
  University of Copenhagen\\
  \texttt{silas@di.ku.dk} \\  
}

\begin{document}
\maketitle

\begin{abstract}
We report on the results of a small crowdsourcing experiment conducted at a workshop on machine learning for segmentation held at the Danish Bio Imaging network meeting 2020. During the workshop we asked participants to manually segment mitochondria in three 2D patches. The aim of the experiment was to illustrate that manual annotations should not be seen as the ground truth, but as a reference standard that is subject to substantial variation. In this note we show how the large variation we observed in the segmentations can be reduced by removing the annotators with worst pair-wise agreement. Having removed the annotators with worst performance, we illustrate that the remaining variance is semantically meaningful and can be exploited to obtain segmentations of cell boundary and cell interior.
\end{abstract}

\keywords{Annotation \and Crowdsourcing \and Mitochondria \and EM}

\section{Introduction}
Manual annotation is the gold standard for many image analysis tasks and a crucial part of developing machine learning methods. Despite well defined annotation protocols and training most tasks will be subject to substantial inter- and intra-annotator variation\cite{joskowicz2019inter,becker2019variability}. Although domain knowledge and experience are important for understanding and interpreting the images, it is not a magic tool that eliminates variation. On the contrary, \cite{joskowicz2019inter} did not find that more experienced radiologists had less variation in annotations. A compelling illustration that lack of familiarity with a task can be helpful is provided by \cite{gurari2016investigating} where simply flipping images improved annotation performance of crowd workers.

We can split variation into different parts: variation due to errors, for example, unintended clicks or annotating the wrong structure; variation due to interpretation of the task, for example, including or excluding edema when asked to segment a brain tumor; and variation due to the inherent ambiguity of the task, for example, partial volume effects and poor signal. The first kind of variation is noise we would like to eliminate completely; the second kind is semantically meaningful information we would like to exploit; and the third kind is a measure of uncertainty regarding the actual true segmentation which we would like to explore.

In this report we illustrate simple approaches for investigating variation, removing noisy annotations and exploiting variation in task interpretation. We hope this note can serve as a reminder that variation in annotations should not be ignored.

\section{Materials \& Methods}
\label{sec:materials-methods}
The crowdsourcing experiment was conducted during an online workshop on machine learning for segmentation held at the Danish Bio Imaging network meeting 2020\footnote{\url{https://www.conferencemanager.dk/danish-bioimaging-meeting-2020/}}. Approximately 45 persons participated in the workshop, which was held online using Zoom\footnote{\url{https://zoom.us/}}. We did not record information regarding the participants scientific background, but assume all have some familiarity with bioimaging. In the experiment we asked participants to segment mitochondria in EM images using an online annotation tool.

\subsection{Data}
\begin{figure}
    \centering
    \includegraphics[width=0.33\textwidth]{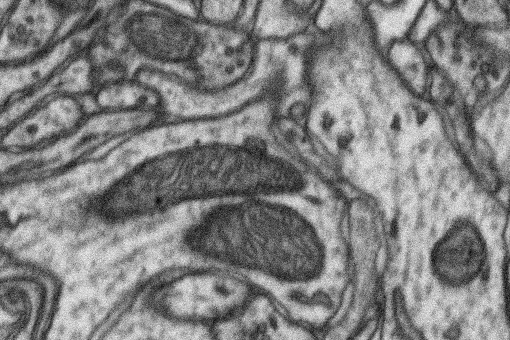}
    \includegraphics[width=0.33\textwidth]{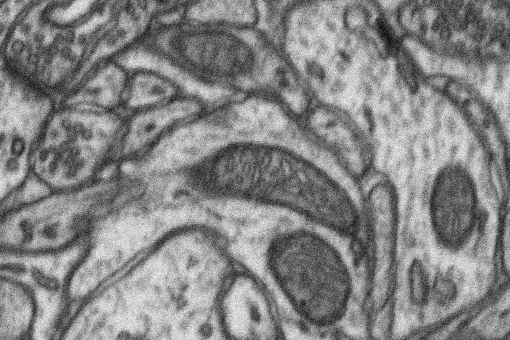}
    \includegraphics[width=0.33\textwidth]{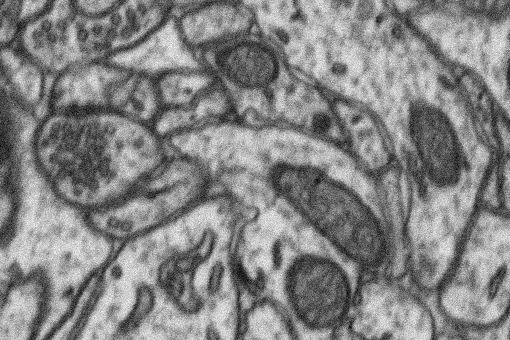}
    \includegraphics[width=0.33\textwidth]{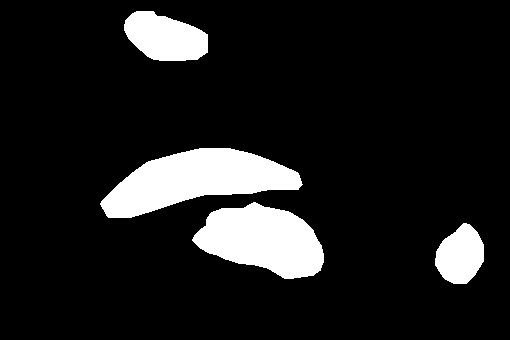}
    \includegraphics[width=0.33\textwidth]{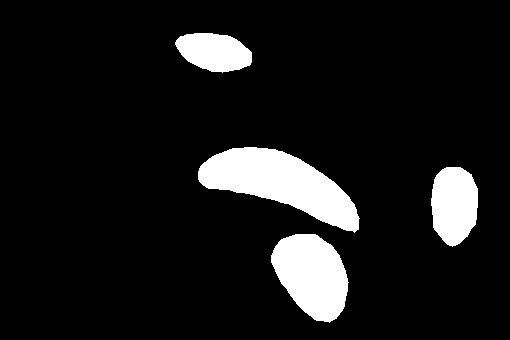}
    \includegraphics[width=0.33\textwidth]{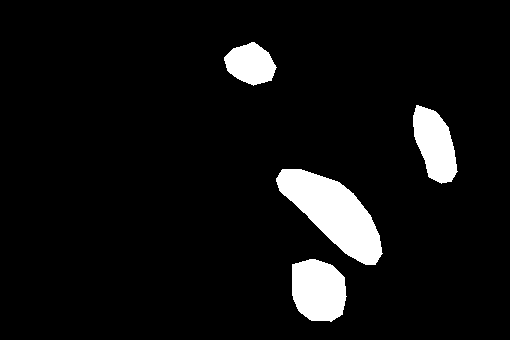}
    \caption{Patches used for crowdsourcing experiment. Top: EM images. Bottom: Reference segmentations}
    \label{fig:data}
\end{figure}

We used three 2D patches extracted from the Electron Microscopy Dataset released by the CVLAB at EPFL\footnote{\url{https://www.epfl.ch/labs/cvlab/data/data-em/}}. The three patches, extracted from the slice 1, 34, and 67 of the training sub-volume, are shown in \autoref{fig:data} alongside reference segmentations from the published dataset.

\subsection{Annotation tools}
\begin{figure}
    \centering
    \includegraphics[width=\textwidth]{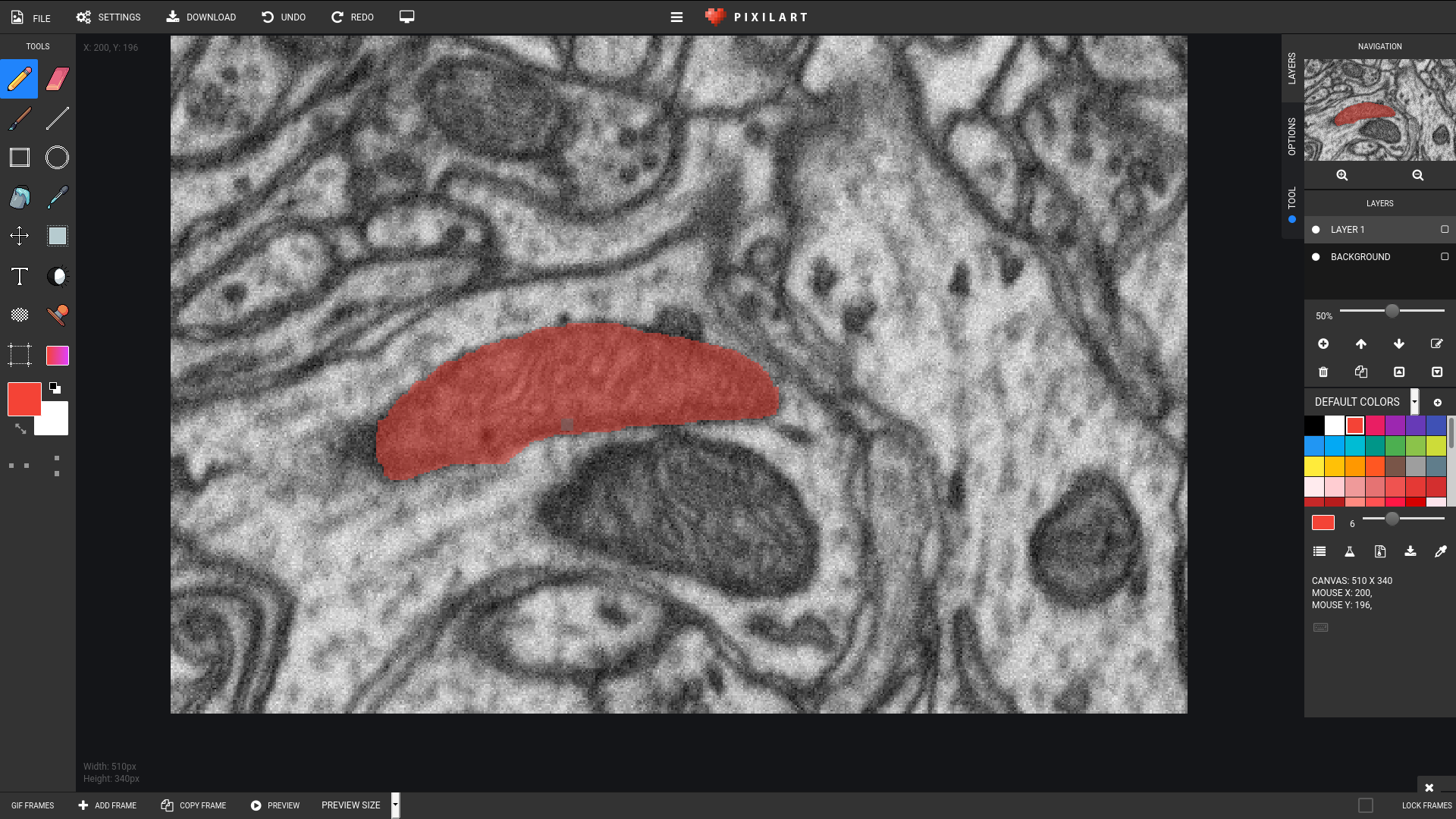}
    \includegraphics[width=\textwidth]{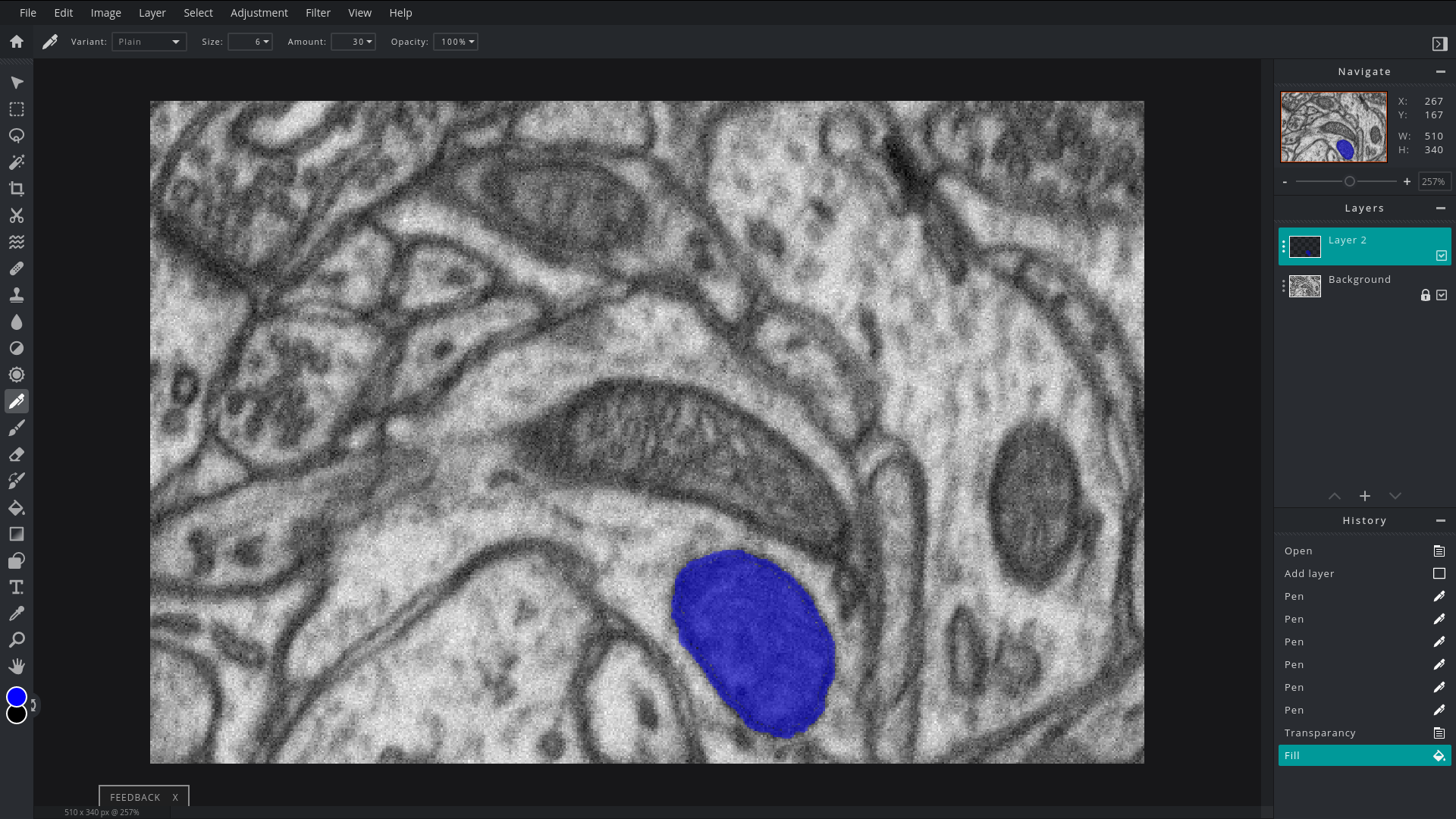}
    \caption{The two annotation tools used for the experiments. Top: Pixilart. Bottom: Pixlr}
    \label{fig:tools}
\end{figure}

We used two online drawing tools for the annotation task, Pixilart\footnote{\url{https://www.pixilart.com/}} and Pixlr\footnote{\url{https://pixlr.com/}}, both illustrated in \autoref{fig:tools}

\subsection{Post-processing}
All annotations where binarized by setting a pixel to 1 if the sum of channels where larger than 0. Six annotations included the background image which was removed prior to binarization. Segmentations from one annotator had speckle noise, which was removed by finding connected components and discarding all components with size of one pixel.

\section{Experiments}
The annotation experiment lasted for about 30 minutes and was initiated after a talk on training a U-Net to segment mitochondria. Participants where given a quick overview of the task and a link to written instructions and data. After the overview, participants where randomly assigned to one of two groups, one using Pixilart and one using Pixlr, using the "breakout rooms" feature of Zoom. Each group was then given a live demonstration of how to perform the annotation task, before working on their own.

\section{Results \& Discussion}
\begin{figure}
    \centering
    \includegraphics[width=0.33\textwidth]{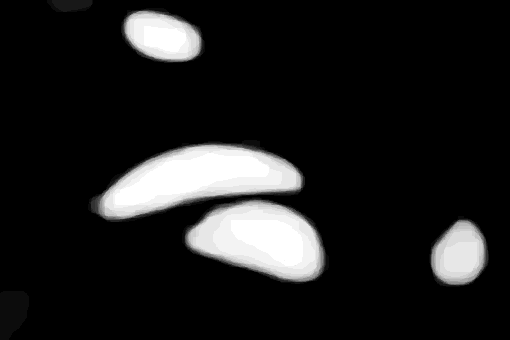}
    \includegraphics[width=0.33\textwidth]{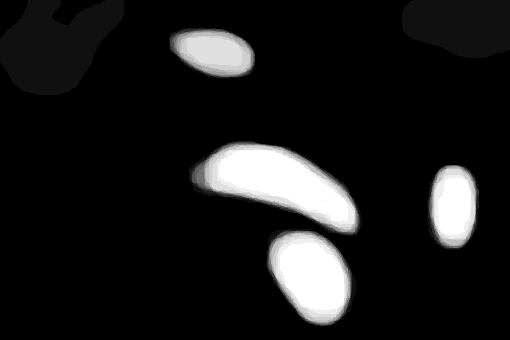}
    \includegraphics[width=0.33\textwidth]{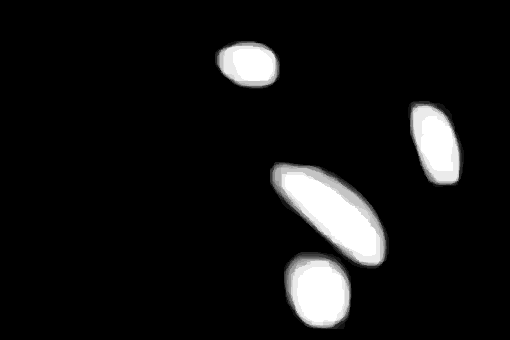}
    \includegraphics[width=0.33\textwidth]{Graphics/0000_training_label.png}
    \includegraphics[width=0.33\textwidth]{Graphics/0033_training_label.png}
    \includegraphics[width=0.33\textwidth]{Graphics/0066_training_label.png}
    \caption{Top: Mean segmentations. Bottom: Reference segmentations. From left to right: Slice 1, 34, 67}
    \label{fig:mean-seg}
\end{figure}

\begin{figure}
    \centering
    \includegraphics[width=0.33\textwidth]{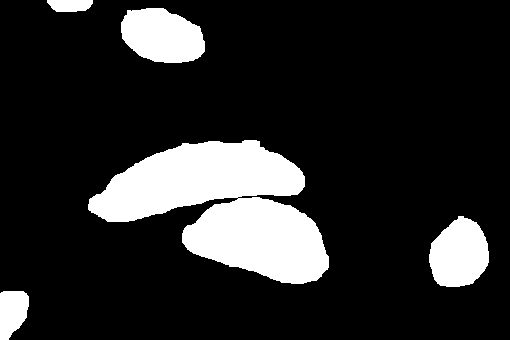}
    \includegraphics[width=0.33\textwidth]{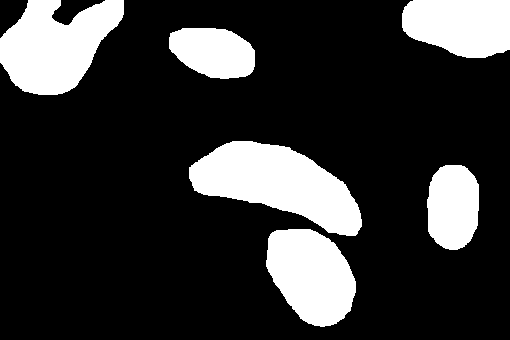}
    \includegraphics[width=0.33\textwidth]{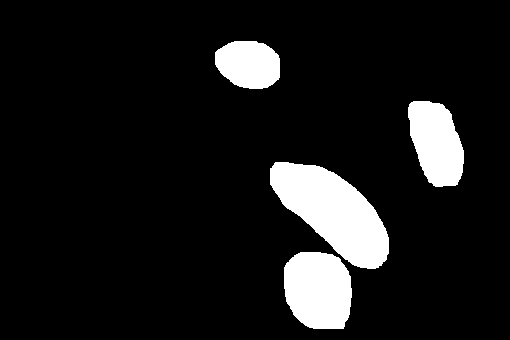}
    \includegraphics[width=0.33\textwidth]{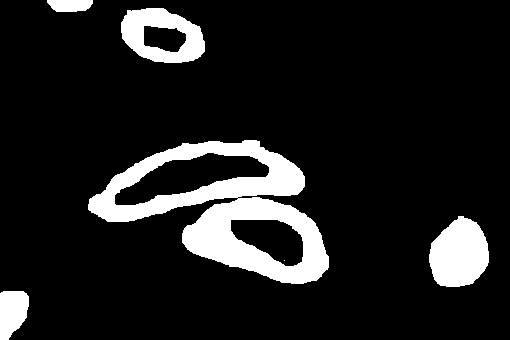}
    \includegraphics[width=0.33\textwidth]{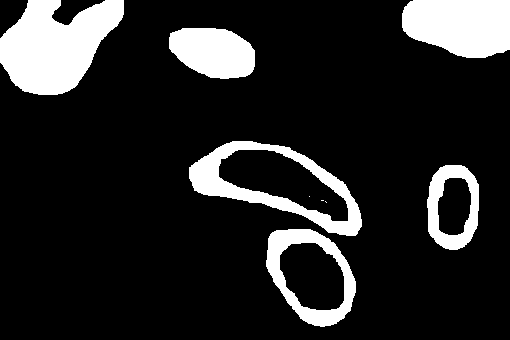}
    \includegraphics[width=0.33\textwidth]{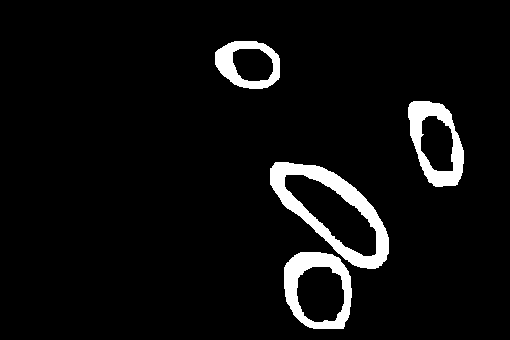}
    \caption{Top: Union of all segmentations. Bottom: Difference between union and intersection of all segmentations. From left to right: Slice 1, 34, 67}
    \label{fig:union-intersection}
\end{figure}

Of the approximately 45 participants, 21 uploaded segmentations of one image, 16 of two images and 13 of all three images. The mean segmentations for each image is shown alongside the reference segmentations in \autoref{fig:mean-seg}. Although the mean segmentations are very close to the references, it is clear that there is substantial variation at the boundary. The extent of this variation is illustrated in \autoref{fig:union-intersection} where the union and (union - intersection) of all segmentations are shown. In the top and middle rows we can see that some annotators missed mitochondria and some added mitochondria. From \autoref{fig:mean-seg} and \autoref{fig:union-intersection} it is clear that at least some variation is due to low quality annotations.

\begin{figure}
    \centering
    \includegraphics[width=0.33\textwidth,trim=55 70 65 81,clip=true]{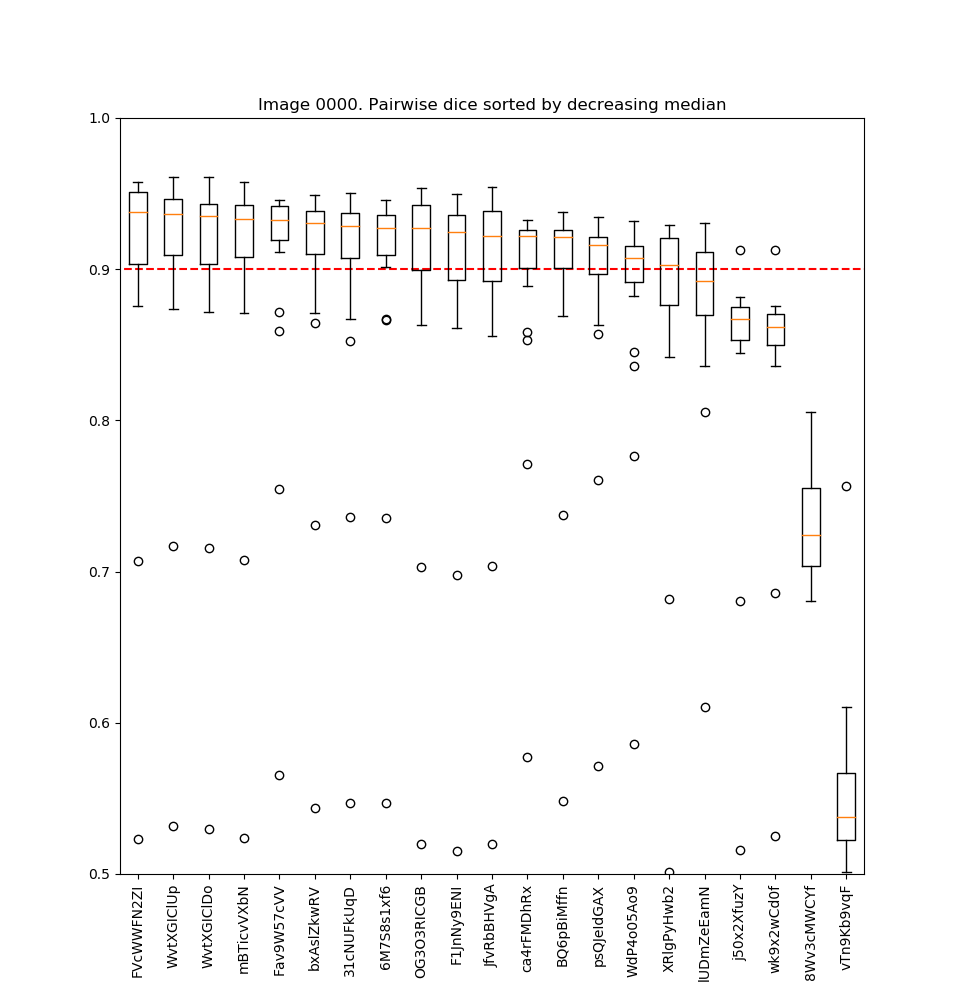}
    \includegraphics[width=0.33\textwidth,trim=55 70 65 81,clip=true]{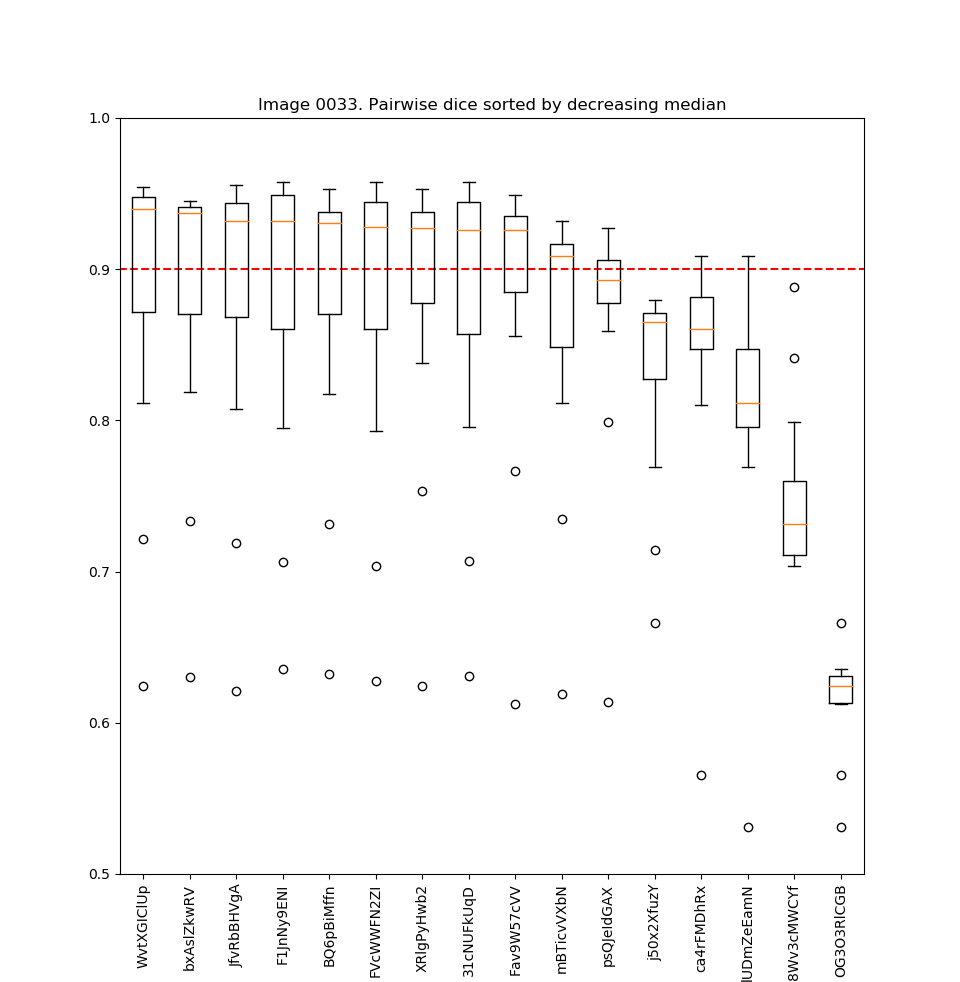}
    \includegraphics[width=0.33\textwidth,trim=55 70 65 81,clip=true]{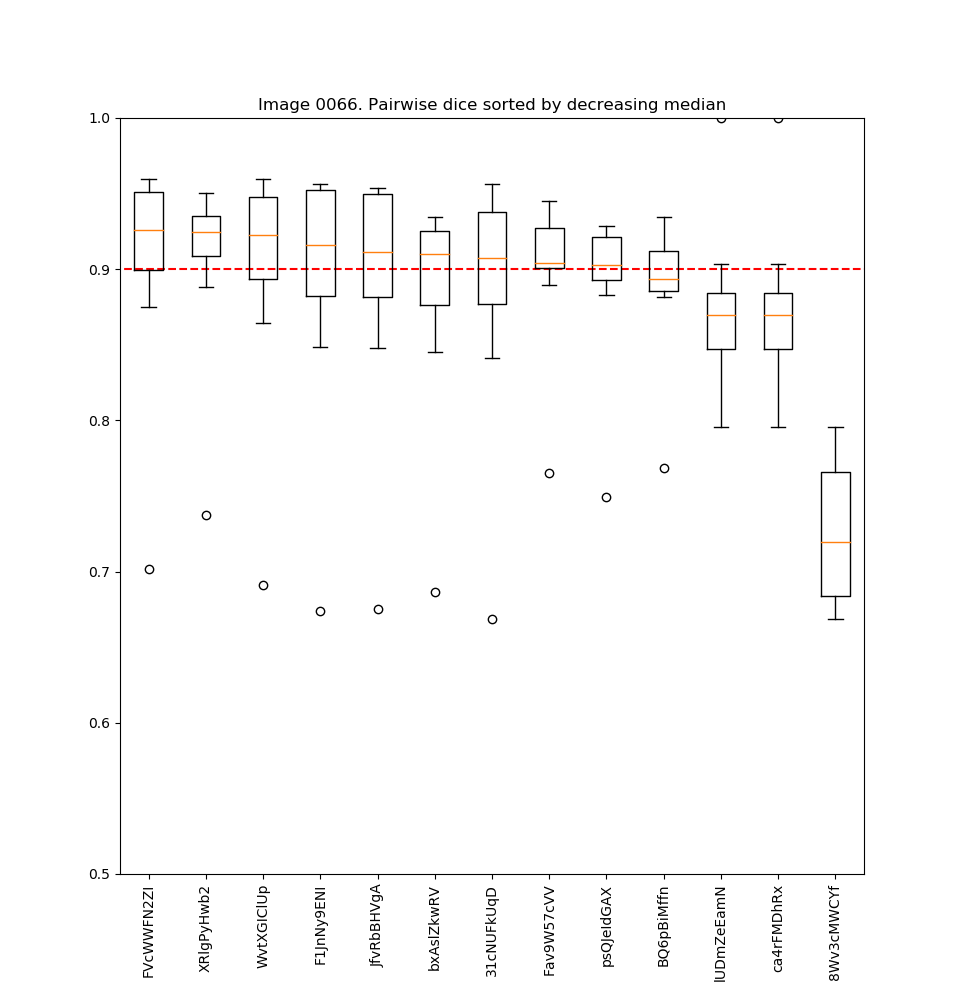}
    \includegraphics[width=0.33\textwidth,trim=55 70 65 81,clip=true]{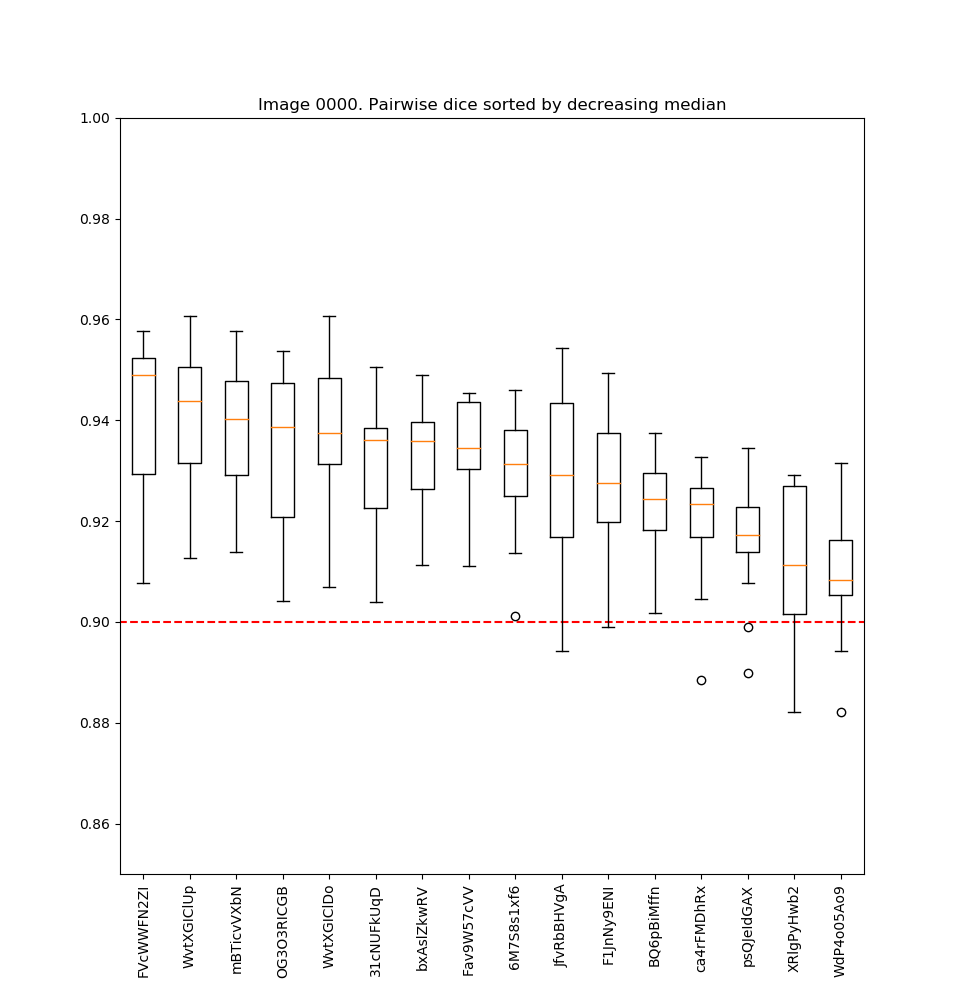}
    \includegraphics[width=0.33\textwidth,trim=55 70 65 81,clip=true]{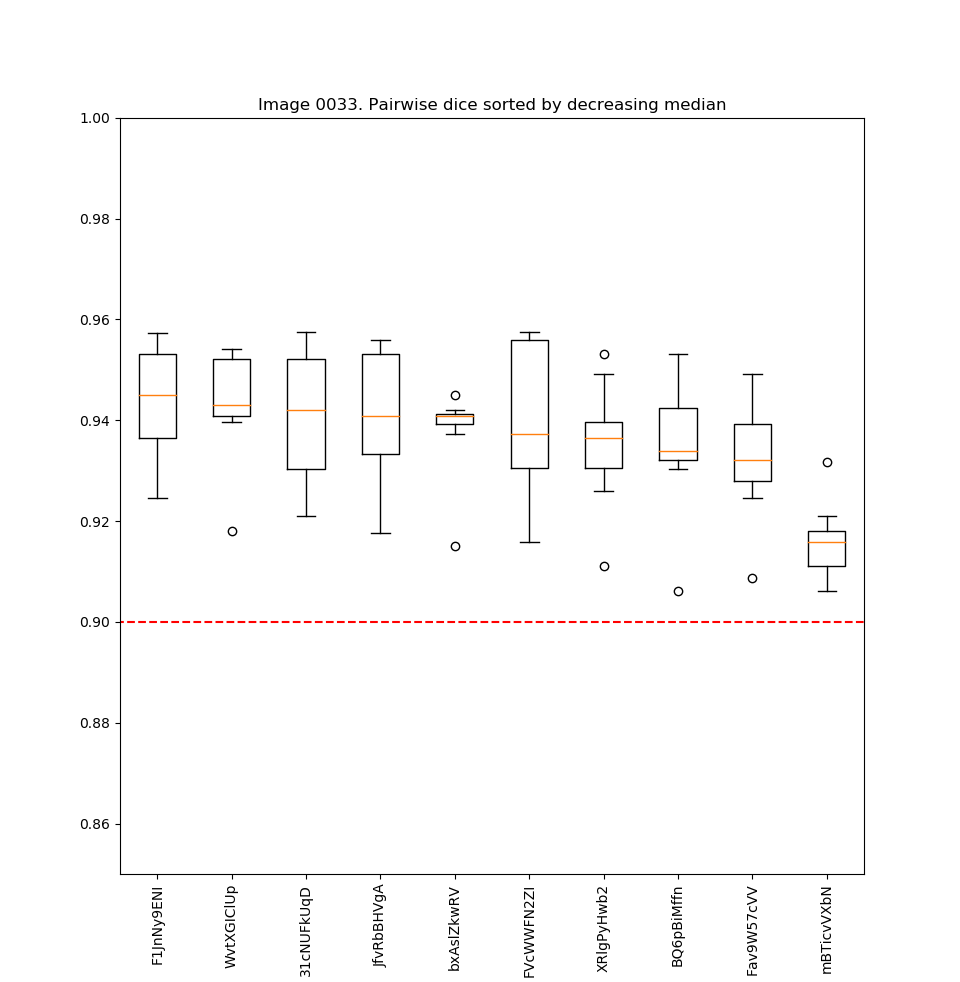}
    \includegraphics[width=0.33\textwidth,trim=55 70 65 81,clip=true]{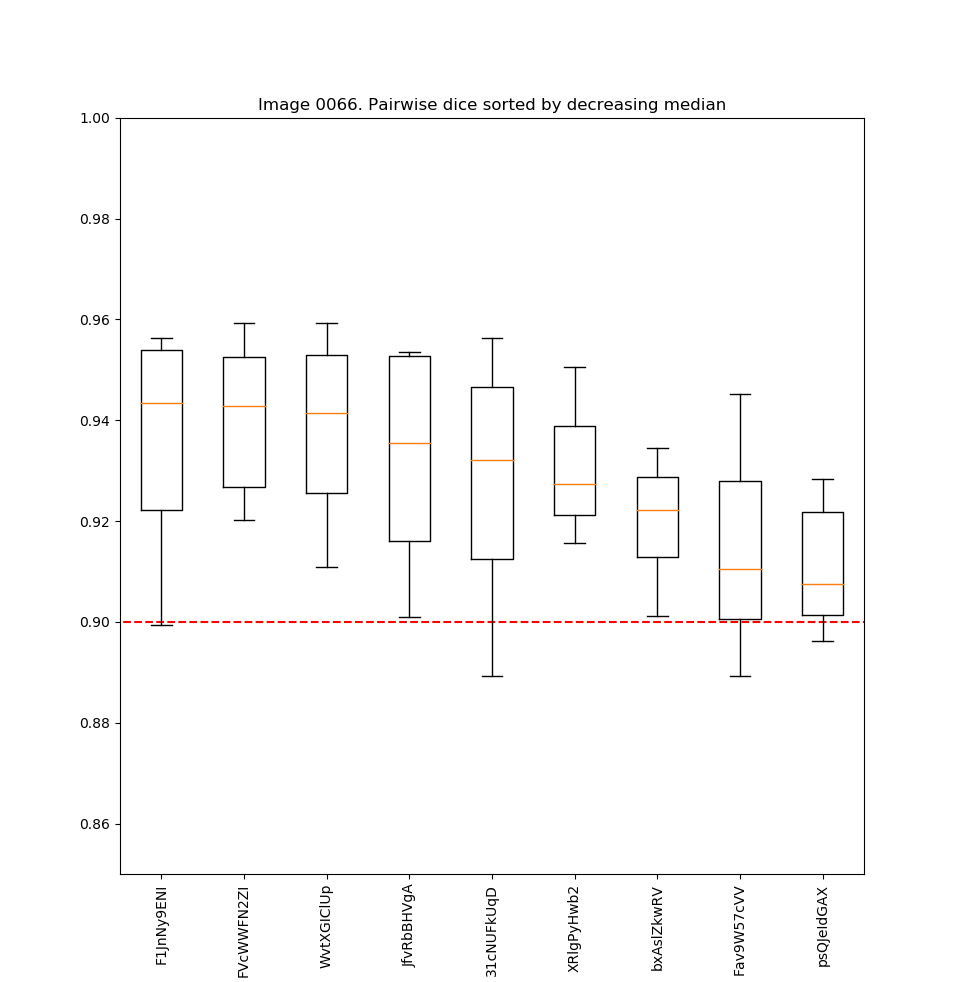}
    \caption{Pairwise dice scores. Dashed red line indicates 0.9 Dice. Top: Before removing annotators. Bottom: After removing annotators. From left to right: Slice 1, 34, 67. }
    \label{fig:pairwise-dice}
\end{figure}

One approach to removing poor annotations is to assume that good annotations will agree with other good annotations, whereas poor annotations will disagree with both good and poor annotations. Since we are dealing with binary segmentations we use the Dice score as measure of agreement and calculate pairwise agreement for all annotators. \autoref{fig:pairwise-dice} shows the distribution of pairwise dice scores for each annotator on each of the three images. The left column shows scores when all annotators are included, the right column shows scores when annotators with median pairwise dice less than 0.9 are excluded. Five annotations where excluded from slice 1, six from slice 34, and four from slice 67.

\begin{figure}
    \centering
    \includegraphics[width=0.33\textwidth]{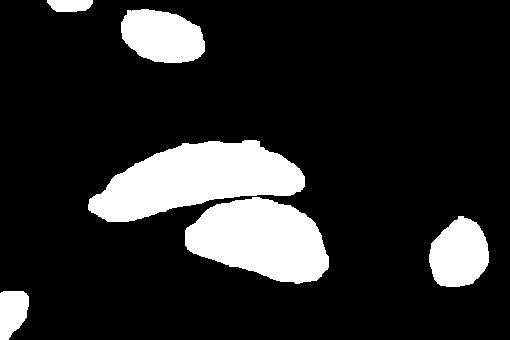}
    \includegraphics[width=0.33\textwidth]{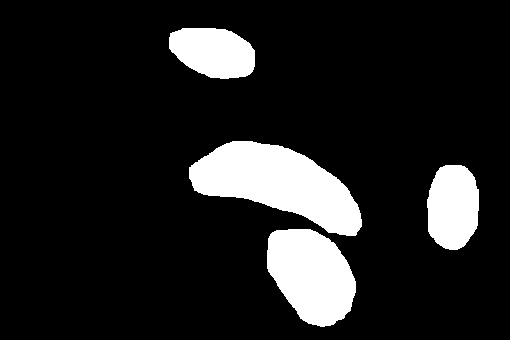}
    \includegraphics[width=0.33\textwidth]{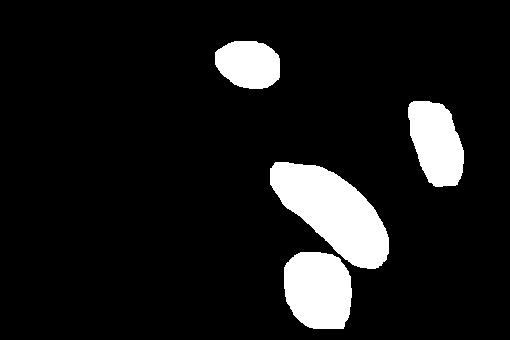}
    \includegraphics[width=0.33\textwidth]{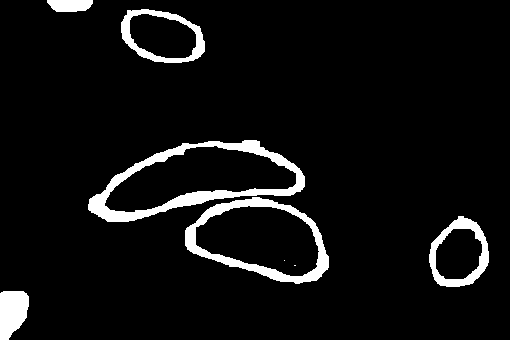}
    \includegraphics[width=0.33\textwidth]{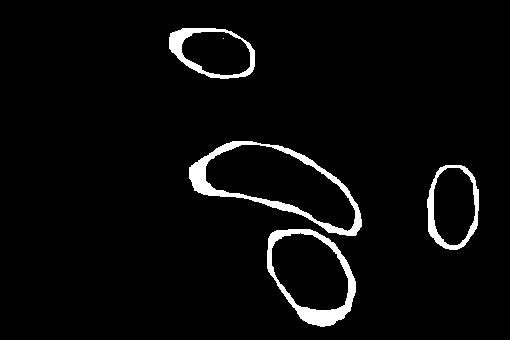}
    \includegraphics[width=0.33\textwidth]{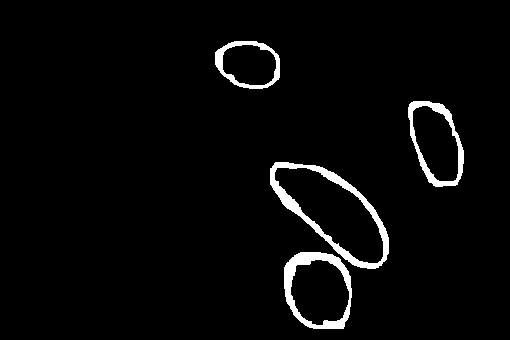}
    \caption{Top: Union of all segmentations. Bottom: Difference between union and intersection of segmentations after removing annotators. From left to right: Slice 1, 34, 67}
    \label{fig:union-intersection-filtered}
\end{figure}

\begin{figure}
    \centering
    \includegraphics[width=0.49\textwidth,trim=150 0 0 0,clip=true]{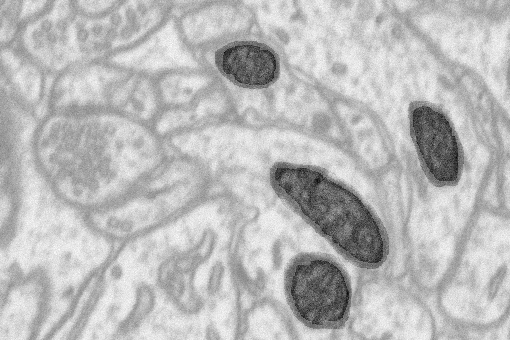}
    \includegraphics[width=0.49\textwidth,trim=150 0 0 0,clip=true]{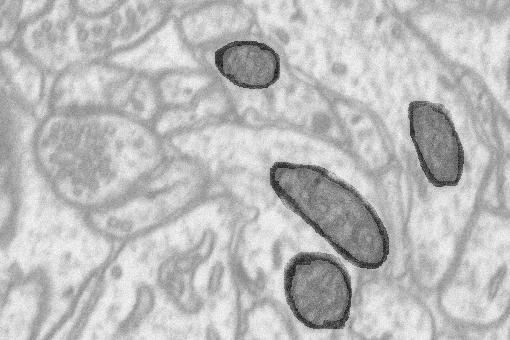}
    \caption{Annotations for slice 67 showing variation can be informative. In this case the area of disagreement corresponds quite well to the boundary of the cell. Left: Intersections. Right: (Union - intersection) }
    \label{fig:information-in-variation}
\end{figure}

The result of excluding annotations with poor agreement is illustrated in \autoref{fig:union-intersection-filtered} where the union and intersection are recomputed for the reduced set of annotations. Results are clearly better with most variation at a comparably small part of the cell boundaries. \autoref{fig:information-in-variation} illustrates that the variation is most likely due to different interpretations of the task, as the part with variation corresponds quite well to the boundary of the cell. Although in general it would be nice to have annotators perform the exact same task, we should be careful treating variation as inherently bad once we have ensured sufficient quality. In this case we get segmentations of the boundary and interior for "free" and can easily get a full cell segmentation as the union.

\printbibliography

\end{document}